\documentclass[conference]{IEEEtran}
\IEEEoverridecommandlockouts
\usepackage{cite}
\usepackage{amsmath,amssymb,amsfonts}
\usepackage{algorithmic}
\usepackage{graphicx}
\usepackage{textcomp}
\usepackage{xcolor}
\usepackage{multirow}
\usepackage{booktabs}
\usepackage{pifont}
\usepackage{caption}

\def\BibTeX{{\rm B\kern-.05em{\sc i\kern-.025em b}\kern-.08em
    T\kern-.1667em\lower.7ex\hbox{E}\kern-.125emX}}
\begin{document}

\title{Coarse-to-fine Alignment Makes Better Speech-image Retrieval\\
\thanks{*Corresponding author.}
}

\author{\IEEEauthorblockN{1\textsuperscript{st} Lifeng Zhou}
\IEEEauthorblockA{\textit{Netease Yidun  AI Lab} \\
Hangzhou, China \\
hzzhoulifeng@corp.netease.com}
\and
\IEEEauthorblockN{2\textsuperscript{nd} Yuke Li\textsuperscript{*}}
\IEEEauthorblockA{\textit{Netease Yidun  AI Lab} \\
Hangzhou, China \\
liyuke@corp.netease.com}
}

\maketitle

\begin{abstract}
In this paper, we propose a novel framework for speech-image
retrieval.
We utilize speech-image contrastive (SIC) learning tasks to align speech and image representations at a coarse level and speech-image
matching (SIM) learning tasks to further refine the fine-grained cross-modal alignment. SIC and SIM learning tasks are jointly trained in a unified manner.
To optimize the learning process, we utilize an
embedding queue that facilitates efficient sampling of high\text{-}quality and diverse negative representations during SIC learning. Additionally, it enhances the learning of SIM tasks by effectively mining hard negatives based on contrastive similarities calculated in SIC tasks. To further optimize learning under noisy supervision, we incorporate momentum distillation into the training process. 
Experimental results show that our framework outperforms the state-of-the-art method by more than 4\% in $\rm{R@1}$ on two benchmark datasets for the speech-image retrieval tasks. Moreover,  as observed in zero-shot experiments, our framework demonstrates excellent generalization capabilities.
\end{abstract}


\section{Introduction}
Speech processing systems have achieved impressive performance by leveraging abundant labeled data and computational resources \cite{Synnaeve_Xu}\cite{Wang_Mohamed_Le_Liu_Xiao_Mahadeokar}. However, the availability of labeled data for most languages is limited, and the process of transcribing large amounts of speech data is costly. Consequently, there has been a growing interest in developing methods that can extract valuable information from unlabeled data \cite{Badino_Canevari_Fadiga_Metta_2014}\cite{Bhati_Nayak_Murty_2017}. Recently, self-supervised learning (SSL) methods have emerged as a prominent approach for learning representations from unlabeled audio data \cite{Synnaeve_Xu}\cite{Schneider_Baevski_Collobert_Auli_2019}\cite{yao2024multimodal}\cite{yao2023improving}. Additionally, exploiting multimodal data and extracting useful information has been explored as another avenue to enhance the performance of speech processing systems. Paired images and speech are extensively used to enhance speech processing, leading to the development of visually grounded speech (VGS) models. 
These models have proven beneficial in various applications, including speech recognition, word discovery, and multilingual spoken language processing. Typically, VGS models are trained and evaluated on speech-image retrieval tasks. 

With the development of VGS models, the accuracy of speech-image retrieval systems has also significantly improved. This showcases that speech-image retrieval holds great appeal as a standalone application. In FaST-VGS \cite{9747103fast-vgs}, authors employ an innovative training and retrieval approach that combines the dual-encoder and cross-attention architectures, resulting in a single model that achieves both speedy and precise speech-image retrieval capabilities. SpeechCLIP \cite{shih2022speechclip} leverages a speech encoder, initially pre-trained with a self-supervised learning (SSL) model, and aligns it with a frozen CLIP \cite{radford2021learning} image encoder using paired speech-image data. This process of aligning speech and image embeddings enables SpeechCLIP to excel in speech-image retrieval tasks, setting new performance benchmarks.

Although these methods have shown effectiveness, they come with certain limitations. For example, in FaST-VGS, using an object detector as the image encoder may limit its expressive power due to the constraints imposed by the detector's capabilities and its predefined visual vocabulary. 
SpeechCLIP substitutes the object detector with the CLIP image encoder to extract image features. However, it primarily relies on contrastive learning tasks to align speech and image features at a high level, which can make achieving precise alignment challenging. This approach can sometimes lead to false positives when the images and speech share similar semantics but differ in finer details.
Moreover, these approaches may be susceptible to noisy data in the training datasets, which can adversely affect their overall generalization performance. 

\begin{figure*}[ht]
  \centering
  \includegraphics[width=\linewidth]{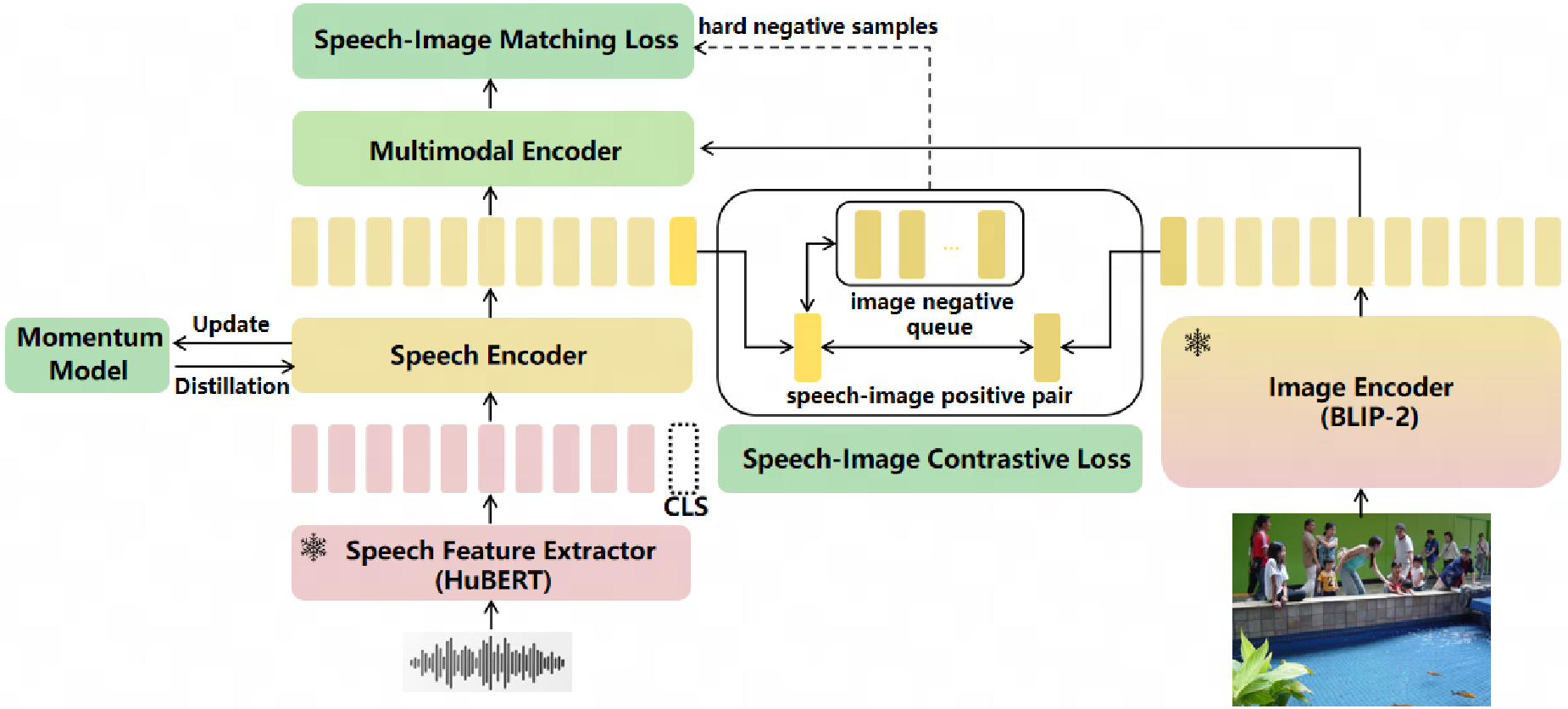}
  \caption{The HuBERT and speech encoder are utilized to extract speech embeddings. The BLIP-2 image encoder is responsible for extracting image embeddings. The speech and image embeddings are fed into the multimodal encoder for interaction.
  We propose SIC and SIM tasks to jointly align speech and image embeddings. 
    We employ a queue that allows for the sampling of diverse negative representations for the SIC tasks and hard negative examples for the SIM tasks.
  In order to improve learning with noisy data, we generate pseudo-targets using the momentum model as additional supervision during training.}
  \label{fig:architecture}
\end{figure*}
Cross-modal alignment is a challenging task, and it is difficult to achieve satisfactory cross-modal alignment solely through a single learning task and simple training scheme, especially when the training data is noisy. In this paper, we utilize multi-task learning and effective training techniques to achieve coarse-to-fine speech-image alignment.
The framework is shown in Figure \ref{fig:architecture}. Our main contributions can be summarized as follows:
\begin{itemize}
\item  We use multitasks speech-image contrastive (SIC) and speech-image matching (SIM) learning tasks
to learn coarse-to-fine alignment between image and speech representations. 
\item  We optimize the learning process by utilizing an embedding queue \cite{he2020momentum}. This queue serves two purposes: firstly, it enables effective sampling of high-quality and diverse negative representations during the SIC learning process. Secondly, it enables efficient sampling hard negative examples for the SIM tasks based on contrastive similarities calculated in SIC tasks, without adding any extra computational overhead.
\item We incorporate momentum distillation \cite{li2021align} into the training process to enhance learning in the presence of noisy data in training datasets. It can be understood as an online self-distillation approach, where the student model learns from a temporal ensemble of itself acting as the teacher model.

\item Our framework has exhibited a significant improvement of over 4\% in $\rm{R@1}$ on the benchmark datasets of Flickr Audio and SpokenCOCO, surpassing the performance of the current state-of-the-art approach.
Furthermore, as observed in the zero-shot experiments, our framework exhibits exceptional generalization capabilities, showcasing its versatility and adaptability across diverse scenarios and datasets.
\end{itemize}.

\section{Method}
\subsection{Preliminaries}
In this section, we will provide a brief explanation of the two pre-trained models, HuBERT and BLIP-2, that are utilized in our framework. 

\noindent\textbf{Hidden-unit BERT (HuBERT)} \cite{hsu2021hubert}.
HuBERT is a self-supervised learning speech model that uses a masked prediction objective, similar to the widely recognized BERT \cite{Devlin_Chang_Lee_Toutanova_2019} model. It predicts masked speech frames by taking into account the surrounding context. This design allows HuBERT to efficiently extract valuable speech representations for a range of downstream tasks \cite{tsai}.

\noindent\textbf{BLIP-2} \cite{Savarese}. BLIP-2 is a highly efficient and adaptable approach for pre-training vision-language models. It leverages frozen pretrained image encoders and large language representation models (LLMs) to align the feature spaces of vision and language, resulting in impressive performance across a range of vision and language tasks. 

\subsection{Architecture}
As shown in Figure \ref{fig:architecture}, our framework comprises a speech feature extractor, two unimodal encoders, and a multimodal encoder. 
We employ a self-supervised learning speech model HuBERT \cite{hsu2021hubert} to extract speech features. 
Taking inspiration from SUPERB \cite{Yang_Chi}, we fuse the CNN output of HuBERT and the hidden representations from its transformer encoder using trainable weights. This weighted aggregation of HuBERT's outputs generates a sequence of frame speech features. These features, along with the CLS token, are then fed into the transformer speech encoder to obtain speech embeddings $S=\{S_{cls}, S_1, ..., S_N\}$.
Additionally, we utilize the visual encoder of BLIP-2 \cite{Savarese} to extract image embeddings $I=\{I_{cls}, I_1, ..., I_N\}$. A transformer multimodal encoder is used to interact between speech and image embeddings. 
We utilize $S_{cls}$ and $I_{cls}$ to facilitate speech-image contrastive learning. Subsequently, image embeddings $I$ and speech embeddings $S$ are then fed to the multimodal encoder for interaction through speech-image matching learning.
To optimize the learning process, we utilize a large image embedding queue that facilitates the incorporation of numerous negative samples during SIC learning. The queue also enables efficient sampling of hard negative examples for the SIM tasks based on contrastive similarities calculated in SIC tasks. Additionally, in order to improve learning with noisy data, we generate pseudo-targets using the momentum model (a moving-average version of the base model) as
additional supervision during training.

\subsection{Training Objectives}
Our model undergoes joint training with two core objectives: conducting speech-image contrastive learning on the unimodal encoders and performing speech-image matching with the multimodal encoder.

\noindent\textbf{Speech-Image Contrastive Learning} is designed to align the speech and the image features at a coarse level, thereby easing the task of the multimodal encoder in cross-modal learning. It optimizes a similarity function $s = S_{cls}^T I_{cls}$, such that matching speech-image pairs receive higher similarity scores than mismatched pairs. Here, $S_{cls}$ and $I_{cls}$ refer to the normalized semantic embeddings of the speech and image, respectively. To effectively increase the diversity of negative examples and enhance the model's ability to discriminate between positive and negative pairs in the contrastive learning process, we utilize a queue to store a fixed size of image embeddings.  
It is worth noting that, unlike in \cite{he2020momentum} where the encoder is updated every iteration, our image encoder remains fixed.
By leveraging the queue, we can access a larger pool of diverse image embeddings for each training iteration. 
For each speech, we calculate the softmax-normalized similarity between the speech and image embeddings as follows:
\begin{equation}
p_j^{\mathrm{s} 2 \mathrm{i}}(S)=\frac{\exp \left(s\left(S, I_j\right) / \tau\right)}{\sum_{j=1}^Q \exp \left(s\left(S, I_j\right) / \tau\right)},
\label{ps2i}
\end{equation}
where $\tau$ is a learnable temperature parameter, $Q$ is the image embedding queue size.
For each image, the softmax-normalized image and speech similarity is calculated as:
\begin{equation}
p_j^{\mathrm{i} 2 \mathrm{s}}(I)=\frac{\exp \left(s\left(I, S_j\right) / \tau\right)}{\sum_{j=1}^B \exp \left(s\left(I, S_j\right) / \tau\right)},
\end{equation}
where $B$ is the batch size. 
Let $\boldsymbol{y}^{\mathrm{s}2\mathrm{i}}(S)$ and $\boldsymbol{y}^{\mathrm{i}2\mathrm{s}}(I)$ represent the ground-truth one-hot similarity, assigning a probability of 0 to negative pairs and 1 to positive pairs. The speech-image contrastive loss is given by the cross-entropy $\mathrm{H}$ between $\boldsymbol{p}$ and $\boldsymbol{y}$, as follows:

\begin{equation}
\mathcal{L}_{\mathrm{sic}}=\frac{1}{2}[\mathrm{H}\left(\boldsymbol{y}^{\mathrm{s} 2 \mathrm{i}}(S), \boldsymbol{p}^{\mathrm{s} 2 \mathrm{i}}(S)\right)\\+\mathrm{H}\left(\boldsymbol{y}^{\mathrm{i} 2 \mathrm{s}}(I), \boldsymbol{p}^{\mathrm{i} 2 \mathrm{s}}(I)\right)]
\end{equation}

The speech-image pairs used for training can be noisy, where positive pairs sometimes are weakly correlated. This means that the speech may contain words that are unrelated to the image, or the image may contain entities that are not mentioned in the speech. Furthermore, in speech-image contrastive learning, negative speeches associated with an image may still contain relevant content. However, the one-hot labels for SIC penalize all negative predictions.
To address these challenges, we propose momentum distillation to force the speech encoder to learn from pseudo-targets generated by a momentum model as shown in Figure \ref{fig:architecture}. The momentum model is the temporal ensemble of the speech encoder.  Its parameters are updated as:
$\theta_{\mathrm{m}} \leftarrow m \theta_{\mathrm{m}}+(1-m) \theta_{\mathrm{s}}$, where $\theta_{\mathrm{s}}$ are the parameters of the speech encoder and $m$ is the momentum coefficient.
During training, we train the predictions of the speech encoder to match that of the momentum model.
Specifically, we first compute the speech-image similarity using features from the momentum model. 
This is done by calculating the dot product between the speech embeddings  $S_{cls}^{'}$ from the momentum model and the image encoder's output $I_{cls}$, denoted as $s^{'} = S_{cls}^{'T} I_{cls}$.
 Then we compute soft pseudo targets  $\boldsymbol{q}^{\mathrm{s} 2 \mathrm{i}}$ by replacing $s$ with $s^{'}$ in Equation \ref{ps2i}. The loss for SIC learning with momentum distillation $\mathcal{L}_{\mathrm{sic}}^{\mathrm{mod}}$ is defined as:
\begin{equation}
\mathcal{L}_{\mathrm{sic}}^{\mathrm{mod}}=(1-\alpha) \mathcal{L}_{\mathrm{sic}}+\alpha \left[\mathrm{KL}\left(\boldsymbol{q}^{\mathrm{s} 2 \mathrm{i}}(S) \| \boldsymbol{p}^{\mathrm{s} 2 \mathrm{i}}(S)\right)\right],
\end{equation}
where $\alpha$ is a balancing factor and $\rm {KL}$ is Kullback-Leibler divergence.

\noindent\textbf{Speech-Image Matching} aims to align the speech and image embeddings at a fine-grained level. It is a binary classification task where the model is asked to predict whether a speech-image pair is positive (matched) or negative (unmatched). To accomplish this, we utilize the output embedding of the CLS token from the multimodal encoder as the joint representation of the speech-image pair. We then pass this representation through a fully connected layer followed by a softmax activation to obtain a two-class probability $p^{sim}$. The SIM loss is:
\begin{equation}
\mathcal{L}_{\mathrm{sim}}= \mathrm{H}\left(\boldsymbol{y}^{\mathrm{sim}}(S,I), \boldsymbol{p}^{\mathrm{sim}}(S,I)\right),
\end{equation}
\begin{table*}[ht]
    \centering
    \setlength{\tabcolsep}{3.42mm}{
    \begin{tabular}{l r r r r r r r r r}
      \toprule
      \multirow{2}{*}{Method} & \multicolumn{3}{c}{ Speech $\rightarrow$ Image } & \multicolumn{3}{c}{ Image $\rightarrow$ Speech } & \multicolumn{3}{c}{ Mean }\\
      \cline{2-4} \cline{5-7} \cline{8-10}
      & R@1 & R@5 & R@10 & R@1 & R@5 & R@10 & R@1 & R@5 & R@10\\
      \hline  & \multicolumn{9}{c}{ Flickr8k } \\
\hline $\rm {FaST\text{-}VGS_{CO}}$ \cite{9747103fast-vgs} & 26.6 & 56.4 & 68.8 & 36.2 & 66.1 & 76.5  & 31.4 & 61.3 & 72.6 \\
\hline $\rm {FaST\text{-}VGS_{CTF}}$ \cite{9747103fast-vgs} & 29.3 & 58.6 & 71.0 & 37.9 & 68.5 & 79.9  & 33.6 & 63.6 & 75.5 \\
\hline MILAN \cite{Sanabria_Waters_Baldridge_2021}& 33.2 & 62.7 & 73.9 & 49.6 & 79.2 & 87.5  & 41.4 & 71.0 & 80.7 \\
\hline Cascaded SpeechCLIP \cite{shih2022speechclip}& 14.7 & 41.2 & 55.1 & 21.8 & 52.0 & 67.7 & 18.3 & 46.6, & 61.4 \\ 
\hline Parallel SpeechCLIP \cite{shih2022speechclip}& 39.1 & 72.0 & 83.0 & 54.5 & 84.5 & 93.2 & 46.8 & 78.3 & 88.1\\
\hline Ours & \textbf{43.8} & \textbf{75.3} & \textbf{85.7} & \textbf{59.0} & \textbf{87.5} & \textbf{95.1} & \textbf{51.4} & \textbf{81.4} & \textbf{90.4} \\ 
\hline & \multicolumn{9}{c}{ SpokenCOCO } \\
\hline ResDAVEne \cite{Hsu_Harwath_Glass_2019} & 17.3 & 41.9 & 55.0 & 22.0 & 50.6 & 65.2 & 19.7 & 46.3 & 60.1\\
\hline $\rm {FaST\text{-}VGS_{CO}}$ \cite{9747103fast-vgs}  & 31.8 & 62.5 & 75.0 & 42.5 & 73.7 & 84.9 & 37.2 & 68.1 & 80.0\\
\hline $\rm {FaST\text{-}VGS_{CTF}}$ \cite{9747103fast-vgs} & 35.9 & 66.3 & 77.9 & 48.8 & 78.2 & 87.0 & 42.4 & 72.3 & 82.5\\
\hline Cascaded SpeechCLIP \cite{shih2022speechclip}& 6.4 & 20.7 & 31.0 & 9.6 & 27.7 & 39.7 & 8.0 & 24.2 & 35.4\\
\hline Parallel SpeechCLIP \cite{shih2022speechclip}& 35.8 & 66.5 & 78.0 & 50.6 & 80.9 & 89.1 & 43.2 & 73.7 & 83.5 \\
\hline Seg. SpeechCLIP \cite{Bhati_Villalba_Moro-Velazquez_Thebaud_Dehak_2023}& 28.2 & 55.3 & 67.5 & 28.5 & 56.1 & 68.9 & 28.4 & 55.7 & 68.2\\
\hline Ours & \textbf{39.9} & \textbf{69.3} & \textbf{80.2} & \textbf{54.9} & \textbf{83.3} & \textbf{90.7} & \textbf{47.4} & \textbf{76.3} & \textbf{85.5}\\
    \bottomrule
    \end{tabular}}
    \caption{Recall scores for speech-image retrieval on Flickr8k and SpokenCOCO testing sets.} 
       \label{tab:main-recall} 
\end{table*}
where $\mathrm{H}$ is cross entropy and $\boldsymbol{y}^{\mathrm{sim}}$ is a 2-dimensional one-hot vector representing the ground-truth label. To improve the model's performance, we propose a strategy to sample hard negatives for the SIM tasks with zero additional computational overhead. A negative speech-image pair is hard if they share similar semantics but differ in fine-grained details. We use the contrastive similarity from Equation \ref{ps2i} which has already been calculated in the SIC tasks to find hard negatives from the image embedding queue. For each speech in a mini-batch, we sample one negative image from the queue. Images with higher contrastive similarities to the speech are more likely to be sampled. 
The full pre-training objective of our framework is donated as:
\begin{equation}
\mathcal{L}=\mathcal{L}_{\mathrm{sic}}^{\mathrm{mod}}+\mathcal{L}_{\mathrm{sim}}
\end{equation}

\section{Experiment}

\subsection{Setup} 
\noindent\textbf{Dataset}. Our model is trained and evaluated with speech-image retrieval on Flickr8k Audio Captions Corpus \cite{Harwath_Glass_2015} and SpokenCOCO dataset \cite{Hsu_Harwath_Miller_Song_Glass_2021}. Each image in both datasets is paired with five spoken captions produced by humans uttering text captions. Flickr8k consists of 8k images and 46 hours of speech, while SpokenCOCO has 123k images and 742 hours of speech. Following FaST-VGS \cite{9747103fast-vgs}, we use the Karpathy \cite{Karpathy_Fei-Fei_2017} split for SpokenCOCO.

\noindent\textbf{Setup}. The Hubert model used in our experiments is Hubert-Large, while the BLIP-2 image encoder is ViT-L/14. Both the HuBERT and BLIP-2 parameters are frozen throughout the training process. The speech encoder and the multimodel encoder are both transformer encoders. They both have eight attention heads, and the hidden dimension of these two encoders is the same as that of HuBERT.  In all our experiments, we set the momentum coefficient $m$ to 0.998 and the balancing factor $\alpha$ to 0.4 for simplicity. The size of the image queue is set differently based on the dataset used for the experiments. The image queue sizes are set to 1024 and 16384 for Flickr8k and SpokenCOCO dataset, respectively. Since the two datasets contain multiple speech for each image, we change the ground-truth label of SIC to consider multiple positives, where each positive has a ground-truth probability of $1/n$, where $n$ is the number of positive samples. During inference, we first compute the feature similarity score $s_{sic}$ for all speech-image pairs. Then we take the top-$k$ candidates and calculate their SIM score $s_{sim}$ for ranking. For the Flickr8k, $k$ is set to 16, while for the SpokenCOCO dataset, $k$ is set to 32. All models are trained with Adam optimizer with a weight decay of $10^{-6}$, batch size of 256, and 40k steps in total. The learning rate linearly increases to $10^{-4}$ in the first 4k steps and decreases to $10^{-8}$ afterward. All experiments are conducted on a machine with 8 32GB V100 GPUs.

\noindent\textbf{Evaluation Metric}.
We select the widely used Recall at K (R@K) metric, where a higher value indicates better performance, to evaluate the cross-modal retrieval performance of our framework. We presented the results for both speech-to-image retrieval and image-to-speech retrieval.

\subsection{Speech-Image Retrieval}
\begin{table*}[ht]
    \centering
    \begin{tabular}{l r r r r r r r r r}
      \toprule
      \multirow{2}{*}{Method} & \multicolumn{3}{c}{ Speech $\rightarrow$ Image } & \multicolumn{3}{c}{ Image $\rightarrow$ Speech } & \multicolumn{3}{c}{ Mean }\\
      \cline{2-4} \cline{5-7} \cline{8-10}
      & R@1 & R@5 & R@10 & R@1 & R@5 & R@10 & R@1 & R@5 & R@10\\
\hline Supervised  & 43.8 & 75.3 & 85.7 & 59.0 & 87.5 & \textbf{95.1} & 51.4 & 81.4 & 90.4 \\
\hline Zero-Shot & \textbf{52.8} & \textbf{81.2} & \textbf{89.8} & \textbf{63.7} & \textbf{88.7} & 93.9  & \textbf{58.3} & \textbf{85.0} & \textbf{91.9} \\ 
    \bottomrule
    \end{tabular}  
    \caption{Recall scores for zero-shot speech-image retrieval on Flickr8k testing sets.} 
    \label{tab:zs-recall} 
\end{table*}

In this section, we evaluate the performance of our framework in the speech-image retrieval tasks, thereby showcasing the effectiveness of our models in aligning speech with image embeddings. As shown in Table \ref{tab:main-recall}, our model surpasses all baseline methods.  
Compared to the result of the previous best model \cite{shih2022speechclip}, our model has achieved significant improvements of 4.2\% in mean $\rm{R@1}$, 3.1\% in mean $\rm{R@5}$, and 2.3\% in mean $\rm{R@10}$ on the Flickr8k dataset. Besides, our model has demonstrated improvements of 4.2\% in mean $\rm{R@1}$, 2.6\% in mean $\rm{R@5}$, and 2.0\% in mean $\rm{R@10}$ on the SpokenCOCO dataset. 
These improvements can be mainly attributed to the ability of our model, jointly trained with SIC and SIM, to not only identify the shared semantics between images and speech but also capture the subtle differences between them.

\begin{table*}[ht] 
    \centering
    \begin{tabular}{c c c r r r r r r r r r r}
      \toprule
      \multicolumn{3}{c}{\multirow{2}{*}{Method}} & \multicolumn{3}{c}{ Speech $\rightarrow$ Image } & \multicolumn{3}{c}{ Image $\rightarrow$ Speech } & \multicolumn{3}{c}{ Mean }\\
      \cline{4-6} \cline{7-9} \cline{10-12}
     & &  & R@1 & R@5 & R@10 & R@1 & R@5 & R@10 & R@1 & R@5 & R@10\\
      \hline Queue & MoD & $\rm {SIM_{hard}}$ & \multicolumn{9}{c}{ Flickr8k } \\
      \hline \ding{55} & \ding{55} & \ding{55}& 39.7 & 72.5 & 83.2 & 55.1 & 85.2 & 93.4 & 47.4 & 78.9 & 88.3\\
      \hline \ding{51} & \ding{55} & \ding{55}&  40.5 & 73.1 & 83.5 & 56.3 & 86.0 & 93.8  & 48.4 & 79.6 & 88.7 \\
      \hline \ding{51} & \ding{51} & \ding{55}  & 41.3 & 73.4 & 84.1 &  56.9 & 86.2 & 94.1  & 49.1 & 79.8 & 89.1 \\
\hline \ding{51} & \ding{51} & \ding{51} & 43.8 & 75.3 & 85.7 & 59.0 & 87.5 & 95.1 & 51.4 & 81.4 & 90.4  \\

\hline  &  &  & \multicolumn{9}{c}{ SpokenCOCO } \\
\hline \ding{55} & \ding{55} & \ding{55} &  36.8 & 66.9 & 78.1 & 51.3 & 81.2& 89.3 & 44.1 & 74.1 & 83.7 \\
\hline \ding{51} & \ding{55} & \ding{55} &  38.1 & 67.4 & 78.4 & 52.5 & 81.5 & 89.6  & 45.3 & 74.5 & 84.0 \\
\hline \ding{51} & \ding{51} & \ding{55} &  38.9 & 67.9 & 78.8 & 53.2 & 81.8 & 89.9  & 46.1 & 74.9 & 84.4 \\
\hline \ding{51} & \ding{51} & \ding{51}& 39.9 & 69.3 & 80.2 & 54.9 & 83.3 & 90.7 & 47.4 & 76.3 & 85.5 \\
    \bottomrule
    \end{tabular}  
    \captionsetup{width=.8\textwidth}
    \caption{Recall scores on Flickr8k and SpokenCOCO testing sets for ablation studies. Queue: image queue. MoD: momentum distillation. $\rm {SIM_{hard}}$: speech-image matching with hard negative mining. In experiments without the $\rm {SIM_{hard}}$ setting, the models are only trained with SIC learning tasks.}
    \label{tab:ab-recall}
\end{table*}
\subsection{Zero-Shot Speech-Image Retrieval}
The generalization capability of a model is a crucial metric. Therefore, in this section, we evaluate the model's generalization by performing zero-shot retrieval. Specifically, we directly assess the model trained on SpokenCOCO using the testing sets of Flickr8K.
The results are presented in Table \ref{tab:zs-recall}, where "Supervised" refers to the model trained on the Flickr8k training sets. Surprisingly, the model trained on the SpokenCOCO training sets achieves significantly better performance than the one trained on the Flickr8k training sets. This underscores the outstanding generalization capability of our model.
This superior performance can be attributed to the larger size of the SpokenCOCO training sets compared to the Flickr8k training sets. We are confident that training on an even larger corpus will further improve its generalization capabilities.

\subsection{Ablation Studies}
In this section, we study the effect of various design choices on speech-image retrieval. The result is shown as Table \ref{tab:ab-recall}. In the experiment without $\rm {SIM_{hard}}$ settings, we exclusively rely on the cosine similarity of the normalized speech and image embeddings for cross-modal retrieval. According to the result, we can conclude that the use of an image embedding queue has facilitated the efficient sampling of diverse negatives during speech-image contrastive learning, resulting in an improvement in the model's performance. Additionally, the application of momentum distillation has contributed to mitigating the influence of noisy data in the training datasets, also boosting the model's performance. Moreover, the integration of $\rm {SIM_{hard}}$ and SIC learning processes has significantly enhanced the model's performance by a substantial margin. We attribute this significant improvement to the effective fusion of embeddings from different modalities in the multi-task learning scheme, which greatly facilitates learning fine-grained cross-modal alignment.

\section{Conclusion}
In this paper, we employ speech-image contrastive and speech-image matching tasks in a joint manner to learn coarse-to-fine alignment between speech and image representations. 
By employing these tasks, our trained model gains the ability to not only identify the shared semantics between images and speech but also capture the subtle differences that exist between them.  Additionally, we incorporate momentum distillation, a form of self-distillation, to help mitigate the impact of noisy data in training datasets.
Furthermore, we employ a large embedding queue to boost the speech-image contrastive and speech-image matching learning process.
With these designs, our framework not only achieves state-of-the-art performance on speech-image retrieval tasks but also exhibits strong generalization and zero-shot ability. These results highlight the effectiveness and robustness of our approach in handling cross-modal retrieval. 
In our future works, we plan to continue making progress towards further boosting its performance, as the accuracy of speech-image retrieval systems has lagged behind their image-text counterparts. Additionally, we aim to investigate the linguistic information learned by the network and transfer our pretrained model to more downstream speech tasks.
\clearpage
\bibliographystyle{IEEEbib}
\bibliography{icme2023template}
\end{document}